\documentclass[11pt,a4paper]{article}
\usepackage[hyperref]{emnlp-ijcnlp-2019}
\usepackage{times}
\usepackage{latexsym}
\usepackage{algorithm}
\usepackage{algorithmic}
\usepackage{amsmath,bm}
\usepackage{url}
\usepackage{hyperref}
\usepackage{helvet}
\usepackage{courier}
\usepackage{colortbl}
\usepackage{multicol}
\usepackage{multirow}
\usepackage{graphicx}
\usepackage{subfigure}
\usepackage{booktabs}       % professional-quality tables
\usepackage{amsfonts}       % blackboard math symbols
\usepackage{nicefrac}       % compact symbols for 1/2, etc.
\usepackage{microtype}      % microtypography
\usepackage{verbatim}
\usepackage{amssymb}% http://ctan.org/pkg/amssymb
\usepackage{pifont}% http://ctan.org/pkg/pifont

\usepackage{wrapfig}
\usepackage{lipsum}
\usepackage{mathrsfs}

\aclfinalcopy % Uncomment this line for the final submission

\newcommand*\samethanks[1][\value{footnote}]{\footnotemark[#1]}

\author{Danqing Wang\thanks{\hspace{1mm} These two authors contributed equally.}, Pengfei Liu\samethanks, Ming Zhong, Jie Fu$\dag$, Xipeng Qiu\thanks{\ \  Corresponding author.}, Xuanjing Huang\\
 School of Computer Science, Fudan University \\
 \&     $\ddag$Mila and Polytechnique Montreal  \\
 \texttt{\{dqwang18,pfliu14,mzhong18,xpqiu,xjhuang\}@fudan.edu.cn} \\
  \texttt{ jie.fu@polymtl.ca}
 }

\newenvironment{itemize*}%
  {\begin{itemize}%
    \setlength{\itemsep}{2pt}%
    \setlength{\parskip}{2pt}}%
  {\end{itemize}}
\newenvironment{enumerate*}%
  {\begin{enumerate}%
    \setlength{\itemsep}{2pt}%
    \setlength{\parskip}{2pt}}%
  {\end{enumerate}}

\newenvironment{enumerate**}%
  {\begin{enumerate}%
    \setlength{\itemsep}{0pt}%
    \setlength{\parskip}{0pt}}%
  {\end{enumerate}}

\title{Exploring Domain Shift in Extractive Text Summarization}

\begin{document}
% controlling equantion space
\setlength{\abovedisplayskip}{2pt}
\setlength{\belowdisplayskip}{2pt}

\maketitle

\begin{abstract}
    Although domain shift has been well explored in many NLP applications, it still has received little attention in the domain of extractive text summarization. As a result, the model is under-utilizing the nature of the training data due to ignoring the difference in the distribution of training sets and shows poor generalization on the unseen domain.
   With the above limitation in mind, in this paper, we first extend the conventional definition of the domain from categories into data sources for the text summarization task. Then we re-purpose a multi-domain summarization dataset and verify how the gap between different domains influences the performance of neural summarization models.
    Furthermore, we investigate four learning strategies and examine their abilities to deal with the domain shift problem.
    Experimental results on three different settings show their different characteristics in our new testbed.

    Our source code including \textit{BERT-based}, \textit{meta-learning} methods for multi-domain summarization learning and the re-purposed dataset \textsc{Multi-SUM} will be available on our project: \url{http://pfliu.com/TransferSum/}.

\end{abstract}

\section{Introduction}
\label{sec:intro}

Text summarization has been an important research topic due to its widespread applications.
Existing research works for summarization mainly revolve around the exploration of neural architectures \cite{cheng2016neural,nallapati2017summarunner} and design of training constraints \cite{paulus2017deep,wu2018learning}. Apart from these, several works try to integrate document characteristics (e.g. domain) to enhance the model performance \cite{haghighi2009exploring,cheung2013probabilistic,cao2017improving,isonuma2017extractive, wang2018reinforced, narayan2018don} or make interpretable analysis towards existing neural summarization models \cite{zhong2019searching}.

Despite their success, only a few literature \cite{cheung2013towards, hua2017pilot} probes into the exact influence domain can bring, while none of them investigates the problem of domain shift, which has been well explored in many other NLP tasks.
This absence poses some challenges for current neural summarization models:
1) How will the domain shift exactly affect the performance of existing neural architectures? 2) How to take better advantage of the domain information to improve the performance for current models? 3) Whenever a new model is built which can perform well on its test set, it should also be employed to unseen domains to make sure that it learns something useful for summarization, instead of overfitting its source domains.

The most important reason for the lack of approaches that deal with domain shift might lay in the unawareness of different domain definitions in text summarization. Most literature limits the concept of the domain into the document categories or latent topics and uses it as the extra loss \cite{cao2017improving,isonuma2017extractive} or feature embeddings \cite{wang2018reinforced, narayan2018don}.
This definition presumes that category information will affect how summaries should be formulated. However, such information may not always be obtained easily and accurately. Among the most popular five summarization datasets, only two of them have this information and only one can be used for training.\footnote{ The five datasets are DUC, Gigward\cite{napoles2012annotated}, CNN/Daily Mail\cite{hermann2015teaching}, The New York Times Annotated Corpus (NYT)\cite{sandhaus2008new} and Newsroom\cite{grusky2018newsroom}. Only DUC and NYT are annotated with document categories, and DUC is designed only for competition test.}
Besides, the semantic categories do not have a clear definition.\footnote{ For example, ``Dining and Wine" in NYT refers to ``Food and Drink" in DUC 2002.} Both of these prevent previous work from the full use of domains in existing datasets or building a new multi-domain dataset that not only can be used for multi-domain learning but also is easy to explore domain connection across datasets.

In this paper, we focus on the extractive summarization and  demonstrate that news publications can cause data distribution differences, which means that they can also be defined as domains. Based on this, we re-purpose a multi-domain summarization dataset \textsc{MULTI-SUM} and further explore the issue of domain shift.

% The property of
Methodologically, we employ four types of models with their characteristics under different settings.
The first model is inspired by the \textit{joint training} strategy, and the second one builds the connection between large-scale pre-trained models and multi-domain learning.
% , in which BERT is used to instruct the processing of domain shift.
The third model directly constructs a domain-aware model by introducing domain type information explicitly.
Lastly, we additionally explore the effectiveness of meta-learning methods to get better generalization. By analyzing their performance under \textsc{in-domain}, \textsc{out-of-domain}, and \textsc{cross-dataset}, we provide a preliminary guideline in Section \ref{sec:quanti} for future research in multi-domain learning of summarization tasks.

Our contributions can be summarized as follows:
\begin{itemize}
    \vspace{-2mm}
    \item We analyze the limitation of the current domain definition in summarization tasks and extend it into article publications. We then re-purpose a dataset \textsc{MULTI-SUM} to provide a sufficient multi-domain testbed (\textsc{in-domain} and \textsc{out-of-domain}). % This requires training and measuring performance on a set of domains instead of solely on single one, encouraging the emergence of systems more robust to domain difference.
    \vspace{-2mm}
    \item To the best of our knowledge, this is the first work that introduces domain shift to text summarization. We also demonstrate how domain shift affects the current system by designing a verification experiment.
    % unsupervised pre-trained models
    \vspace{-2mm}
    \item Instead of pursuing a unified model, we aim to analyze how different choices of model designs influence the generalization ability of dealing with the domain shift problem, shedding light on the practical challenges and provide a set of guidelines for future researchers. % @@@@@@@@@@@@@@@@@@@@@@@@@@@@----------------

\end{itemize}

% Table generated by Excel2LaTeX from sheet 'New_publication'
\begin{table*}[htbp]\small\setlength{\tabcolsep}{6pt}
\centerline{
 \begin{tabular}{lrrrccccccccc}
 \toprule
  & \multicolumn{3}{c}{\textbf{Statistics}} & \multicolumn{3}{c}{\textbf{Measures}} & \multicolumn{3}{c}{\textbf{Lead}} & \multicolumn{3}{c}{\textbf{Ext-Oralce}} \\
 \midrule
  & \multicolumn{1}{c}{\textbf{Train}} & \multicolumn{1}{c}{\textbf{Valid}} & \multicolumn{1}{c}{\textbf{Test}} & \multicolumn{1}{c}{\textbf{Cov.}} & \multicolumn{1}{c}{\textbf{Den.}} & \multicolumn{1}{c}{\textbf{Comp.}} & \multicolumn{1}{c}{\textbf{R-1}} & \multicolumn{1}{c}{\textbf{R-2}} & \multicolumn{1}{c}{\textbf{R-L}} & \multicolumn{1}{c}{\textbf{R-1}} & \multicolumn{1}{c}{\textbf{R-2}} & \multicolumn{1}{c}{\textbf{R-L}} \\
 \midrule
 FN & \cellcolor[rgb]{.851,.851,.851} 78,760  & \cellcolor[rgb]{.851,.851,.851}  8,423  & \cellcolor[rgb]{.851,.851,.851}  8,392  & 0.90  & 16.18 & 35.58 & \cellcolor[rgb]{.851,.851,.851}40.30 & \cellcolor[rgb]{.851,.851,.851}33.90 & \cellcolor[rgb]{.851,.851,.851}38.74 & 73.61 & 65.53 & 71.50 \\
 CNN & \cellcolor[rgb]{.851,.851,.851} 43,466  & \cellcolor[rgb]{.851,.851,.851}  4,563  & \cellcolor[rgb]{.851,.851,.851}  4,619  & 0.85  & 12.46 & 38.28 & \cellcolor[rgb]{.851,.851,.851}35.56 & \cellcolor[rgb]{.851,.851,.851}25.60 & \cellcolor[rgb]{.851,.851,.851}33.25 & 59.99 & 46.66 & 56.64 \\
 MA & \cellcolor[rgb]{.851,.851,.851} 31,896  & \cellcolor[rgb]{.851,.851,.851}  3,414  & \cellcolor[rgb]{.851,.851,.851}  3,316  & 0.84  & 6.66  & 28.93 & \cellcolor[rgb]{.851,.851,.851}29.38 & \cellcolor[rgb]{.851,.851,.851}19.15 & \cellcolor[rgb]{.851,.851,.851}27.17 & 55.35 & 40.97 & 51.97 \\
 NYT & \cellcolor[rgb]{.851,.851,.851}  152,959  & \cellcolor[rgb]{.851,.851,.851} 16,488  & \cellcolor[rgb]{.851,.851,.851} 16,620  & 0.85  & 9.19  & 42.30 & \cellcolor[rgb]{.851,.851,.851}28.24 & \cellcolor[rgb]{.851,.851,.851}16.62 & \cellcolor[rgb]{.851,.851,.851}25.20 & 52.25 & 36.14 & 47.73 \\
 WTP & \cellcolor[rgb]{.851,.851,.851} 95,379  & \cellcolor[rgb]{.851,.851,.851}  9,939  & \cellcolor[rgb]{.851,.851,.851} 10,072  & 0.76  & 6.04  & 63.52 & \cellcolor[rgb]{.851,.851,.851}20.75 & \cellcolor[rgb]{.851,.851,.851}10.57 & \cellcolor[rgb]{.851,.851,.851}18.56 & 43.00 & 27.14 & 39.48 \\
 \midrule
 \textbf{Avg} & \cellcolor[rgb]{.851,.851,.851} 80,492 & \cellcolor[rgb]{.851,.851,.851} 8,565 & \cellcolor[rgb]{.851,.851,.851} 8,604  & 0.84  & 10.11 & 41.72 & \cellcolor[rgb]{.851,.851,.851}29.70 & \cellcolor[rgb]{.851,.851,.851}19.74 & \cellcolor[rgb]{.851,.851,.851}27.30 & 55.32 & 41.28 & 51.72 \\
 \midrule
 NYDN  & \cellcolor[rgb]{.851,.851,.851} 55,653  & \cellcolor[rgb]{.851,.851,.851}  6,057  & \cellcolor[rgb]{.851,.851,.851}  5,904  & 0.93  & 14.57 & 21.25 & \cellcolor[rgb]{.851,.851,.851}45.25 & \cellcolor[rgb]{.851,.851,.851}37.69 & \cellcolor[rgb]{.851,.851,.851}43.64 & 74.05 & 64.84 & 72.13 \\
 WSJ & \cellcolor[rgb]{.851,.851,.851} 49,967  & \cellcolor[rgb]{.851,.851,.851}  5,449  & \cellcolor[rgb]{.851,.851,.851}  5,462  & 0.80  & 8.45  & 23.64 & \cellcolor[rgb]{.851,.851,.851}35.21 & \cellcolor[rgb]{.851,.851,.851}23.70 & \cellcolor[rgb]{.851,.851,.851}32.26 & 57.21 & 43.08 & 53.31 \\
 USAT  & \cellcolor[rgb]{.851,.851,.851} 44,919  & \cellcolor[rgb]{.851,.851,.851}  4,628  & \cellcolor[rgb]{.851,.851,.851}  4,781  & 0.78  & 6.35  & 31.17 & \cellcolor[rgb]{.851,.851,.851}25.11 & \cellcolor[rgb]{.851,.851,.851}15.52 & \cellcolor[rgb]{.851,.851,.851}23.03 & 47.22 & 33.43 & 44.05 \\
 TG & \cellcolor[rgb]{.851,.851,.851} 58,057  & \cellcolor[rgb]{.851,.851,.851}  6,376  & \cellcolor[rgb]{.851,.851,.851}  6,273  & 0.80  & 2.75  & 40.35 & \cellcolor[rgb]{.851,.851,.851}21.66 & \cellcolor[rgb]{.851,.851,.851}8.02 & \cellcolor[rgb]{.851,.851,.851}18.24 & 41.23 & 21.56 & 35.90 \\
 TIME  & \cellcolor[rgb]{.851,.851,.851} 42,200  & \cellcolor[rgb]{.851,.851,.851}  4,761  & \cellcolor[rgb]{.851,.851,.851}  4,702  & 0.75  & 4.87  & 47.67 & \cellcolor[rgb]{.851,.851,.851}19.80 & \cellcolor[rgb]{.851,.851,.851}10.83 & \cellcolor[rgb]{.851,.851,.851}17.94 & 41.37 & 26.04 & 37.87 \\
 \midrule
 \textbf{Avg} & \cellcolor[rgb]{.851,.851,.851} 50,159 & \cellcolor[rgb]{.851,.851,.851} 5,454  & \cellcolor[rgb]{.851,.851,.851} 5,424  & 0.81  & 7.40  & 32.81 & \cellcolor[rgb]{.851,.851,.851}29.90 & \cellcolor[rgb]{.851,.851,.851}19.54 & \cellcolor[rgb]{.851,.851,.851}27.48 & 52.79 & 38.33 & 49.20 \\
 \bottomrule
 \end{tabular}}%
 \caption{The statistics of the \textsc{MULTI-SUM} dataset. Three measures refer to \textsc{Coverage}, \textsc{Density} and \textsc{Compression} respectively. \textsc{Lead} and \textsc{Ext-Oracle} are two common baselines for summarization. All measures and baselines are calculated on the test set of the corresponding publication. The top five publication are used as source domains for training and the bottom ones are viewed as \textsc{out-of-domain}.}
  \label{tab:twosubsets}%
\end{table*}%

\vspace{-5pt}

\section{Domains in Text Summarization}
\label{sec:domainshift}

In this section, we first describe similar concepts used as the domain in summarization tasks. Then we extend the definition into article sources and verify its rationality through several indicators that illustrate the data distribution on our re-purposed multi-domain summarization dataset.

\subsection{Common Domain Definition}
% 尽管domain常常被按照文本或者图片的内容类别定义，然而domain最初引入的动机是想描述数据集中影响样本分布不同的因素，因为他会给模型的学习带来挑战。
Although a domain is often defined by the content category of a text \cite{li2008multi,blitzer2007biographies} or image \cite{saenko2010adapting}, the initial motivation for a domain is a metadata attribute which is used in order to divide the data into parts with different distributions \cite{joshi2012multi}.

For text summarization, the differences between data distribution are often attributed to the document categories, such as sports or business, or the latent topics within articles, which can be caught by classical topic models like Latent Dirichlet Allocation (LDA) \cite{blei2003latent}. Although previous works have shown that taking consideration of those distribution differences can improve summarization models performance \cite{isonuma2017extractive, wang2018reinforced}, few related them with the concept of the domain and investigated the summarization tasks from a perspective of multi-domain learning. \footnote{ \citet{hua2017pilot} studied domain adaptation between news stories and opinion articles from NYT. However, their model was just trained in a single domain and was adapted to another, which was different from our multi-domain training and evaluation settings.}

\subsection{Publications as Domain}
% @@@@@@@@@@@@@@@@@@@@@@@@@@@@----------------
In this paper, we extend the concept into the article sources, which can be easily obtained and clearly defined\footnote{ Most existing benchmark datasets are a mixture of multiple publications with the idea of collecting a larger amount of data, such as CNN/DailyMail, Gigward and Newsroom.}.

% \vspace{-5pt}
\paragraph{Three Measures} We assume that the publications of news may also affect data distribution and thus influence the summarization styles. In order to verify our hypothesis, we make use of three indicators (\textsc{Coverage}, \textsc{Density} and \textsc{Compression}) defined by \citet{grusky2018newsroom} to measure the overlap and compression between the (document, summary) pair. The coverage and the density are the word and the longest common subsequence (LCS) overlaps, respectively. The compression is the length ratio between the document and the summary.

% \vspace{-5pt}
\paragraph{Two Baselines} We also calculate two strong summarization baselines for each publication. The \textsc{LEAD} baseline concatenates the first few sentences as the summary and calculates its ROUGE score. This baseline shows the lead bias of the dataset, which is an essential factor in news articles. The \textsc{Ext-Oracle} baseline evaluates the performance of the ground truth labels and can be viewed as the upper bound of the extractive summarization models \cite{nallapati2017summarunner, narayan2018don}.

% \vspace{-5pt}
\paragraph{\textsc{MULTI-SUM}} The recently proposed dataset Newsroom \cite{grusky2018newsroom} is used, which was scraped from 38 major news publications. We select top ten publications (\textit{NYTimes}, \textit{WashingtonPost}, \textit{FoxNews}, \textit{TheGuardian}, \textit{NYDailyNews}, \textit{WSJ}, \textit{USAToday}, \textit{CNN}, \textit{Time} and \textit{Mashable}) and process them in the way of \citet{See2017}. To obtain the ground truth labels for extractive summarization task, we follow the greedy approach introduced by \citet{nallapati2017summarunner}. Finally, we randomly divide ten domains into two groups, one for training and the other for test. We call this re-purposed subset of Newsroom \textsc{MULTI-SUM} to indicate it is specially designed for multi-domain learning in summarization tasks.

% \vspace{-10pt}
From Table \ref{tab:twosubsets}, we can find that data from those news publications vary in indicators that are closely relevant to summarization. This means that (document, summary) pairs from different publications will have unique summarization formation, and models might need to learn different semantic features for different publications. Furthermore, we follow the simple experiment by \citet{torralba2011unbiased} to train a classifier for the top five domains. A simple classification model with GloVe initializing words can also achieve 74.84\% accuracy (the chance is 20\%), which ensures us that there is a built-in bias in each publication. Therefore, it is reasonable to view one publication as a domain and use our multi-publication \textsc{MULTI-SUM} as a multi-domain dataset.

\section{Analytical Experiment for Domain Shift}
Domain shift refers to the phenomenon that a model trained on one domain performs poorly on a different domain\cite{saenko2010adapting,gopalan2011domain}. To clearly verify the existence of domain shift in the text summarization, we design a simple experiment on \textsc{MULTI-SUM} dataset.

Concretely, we take turns choosing one domain and use its training data to train the basic model. Then, we use the testing data of the remaining domains to evaluate the model with the automatic metric ROUGE \cite{lin2003automatic}
% and results are reported in Table \ref{tab:results_cross_domains}.

%xpqiu a-j可以换成domain name
% Table generated by Excel2LaTeX from sheet 'domain-shift'
\begin{table*}[htbp]\footnotesize
  \centering
  \begin{tabular}{lrrrrrrrrrr}
  \toprule
      & \multicolumn{1}{c}{FN} & \multicolumn{1}{c}{CNN} & \multicolumn{1}{c}{MA} & \multicolumn{1}{c}{NYT} & \multicolumn{1}{c}{WTP} & \multicolumn{1}{c}{NYDN} & \multicolumn{1}{c}{WSJ} & \multicolumn{1}{c}{USAT} & \multicolumn{1}{c}{TG} & \multicolumn{1}{c}{TIME} \\
  \midrule
    FN & \cellcolor[rgb]{ .816,  .808,  .808} 48.84  &  -1.23  &  -1.76  &  -0.70  &  -0.27  &  -2.29  &  -0.31  &  -0.57  &  -0.27  &  -0.02  \\
    CNN   &  -0.92  &  \cellcolor[rgb]{ .816,  .808,  .808} 41.22  &  -1.93  &  -1.49  &  -1.01  &  -3.83  &  -0.55  &  -1.00  &  -0.80  &  -0.06  \\
    MA & -2.59  &  -6.62  & \cellcolor[rgb]{ .816,  .808,  .808} 35.19  &  -1.37  &  -1.45  &  -5.11  &  -0.25  &  -1.54  &  -0.17  &  -0.71  \\
    NYT &  -3.13  &  -4.46  &  -2.41  & \cellcolor[rgb]{ .816,  .808,  .808} 29.65  &  -0.97  &  -3.95  &  -0.33  &  -1.16  &  -0.17  &  -0.86  \\
    WTP &  -1.92  &  -3.03  &  -2.00  &  -0.61  & \cellcolor[rgb]{ .816,  .808,  .808} 23.01  &  -3.28  &  -0.43  &  -0.57  &  -0.01  &  -0.28  \\
    NYDN &  -1.96  &  -2.31  &  -1.91  &  -0.57  &  -0.53  & \cellcolor[rgb]{ .816,  .808,  .808} 51.36  &  -0.41  &  -1.20  &  -0.36  &  -0.05  \\
    WSJ   &  -4.66  &  -7.07  &  -3.10  &  -1.04  &  -1.57  &  -6.37  & \cellcolor[rgb]{ .816,  .808,  .808} 38.60  &  -1.71  &  -0.58  &  -1.00  \\
    USAT &  -2.04  &  -3.55  &  -2.56  &  -2.77  &  -1.86  &  -6.09  &  -0.92  & \cellcolor[rgb]{ .816,  .808,  .808} 29.11  &  -0.68  &  -1.27  \\
    TG &  -5.47  &  -8.63  &  -3.50  &  -1.50  &  -1.74  &  -5.82  &  -1.43  &  -2.19  & \cellcolor[rgb]{ .816,  .808,  .808} 23.93  &  -1.66  \\
    TIME  &  -0.82  &  -3.05  &  -1.64  &  -1.36  &  -1.06  &  -4.65  &  -0.47  &  -1.05  &  -0.24  & \cellcolor[rgb]{ .816,  .808,  .808} 21.90  \\
  \bottomrule
  \end{tabular}%
  \caption{Results (Matrix $V$) of the verification experiment based on the \textsc{MULTI-SUM} dataset.
  The ROUGE-1 scores \protect\footnotemark{} of the model which is trained and tested on the same domain $R_{ii}$ are shown on the diagonal line. It is regarded as benchmark scores. The other cells $V_{ij} = R_{ij}-R_{jj}, i \neq j$, which represents that for the same test domain $j$, how many improvements we obtained when we switch from training domain $i$ to $j$.
  Positive values are higher than the benchmark, and negative values are less than the benchmark.}
  \label{tab:results_cross_domains}%
\end{table*}%
\footnotetext{ROUGE-2 and ROUGE-L show similar trends and their results are attached in Appendix.}

% \vspace{-5pt}
\paragraph{Basic Model}
\label{sec:basicmodel}
Like a few recent approaches, we define extractive summarization as a sequence labeling task. %\cite{Cheng2016, nallapati2017summarunner, Isonuma2017empirical, Yasunaga2017, Narayan2018rank, narayan2018don}
Formally, given a document $S$ consisting of $n$ sentences $s_1, \cdots, s_n$, the summaries are extracted by predicting a sequence of label $Y = y_1, \cdots, y_n$ ($y_i \in \{0,1\}$) for the document, where $y_i = 1$ represents the $i$-th sentence in the document should be included in the summaries.

% \footnote{We test our basic model performance on the non-anonymized CNN/Daily Mail dataset and the whole Newsroom. The results are presented in Appendix.}
In this paper, we implement a simple but powerful model based on the encoder-decoder architecture. We choose CNN as the sentence encoder following prior works \cite{Chen2018fast} and employ the popular modular Transformer \cite{Vaswani2017} as the document encoder. The detailed settings are described in Section \ref{sec:impl}.

% \vspace{-5pt}
%\subsection{Poor Cross-domain Generalization}
\paragraph{Results}

From Table \ref{tab:results_cross_domains}, we find that the values are negative except the diagonal, which indicates models trained and tested on the same domain show the great advantage to those trained on other domains. The significant performance drops demonstrate that the domain shift problem is quite serious in extractive summarization tasks, and thus pose challenges to current well-performed models, which are trained and evaluated particularly under the strong hypothesis: training and test data instances are drawn from the identical data distribution. Motivated by this vulnerability, we investigate the domain shift problem under both multi-domain training and evaluation settings.

\section{Multi-domain Summarization}
\label{sec:models}
With the above observations in mind, we are seeking an approach which can alleviate the domain shift problem effectively in text summarization. Specifically, the model should not only perform well on source domains where it is trained on, but also show advantage on the unseen target domains. This involves the tasks of multi-domain learning and domain adaptation. Here, we begin with several simple approaches for multi-domain summarization based on multi-domain learning.

\subsection{Four Learning Strategies}
To facilitate the following description, we first set up mathematical notations.
Assuming that there are $K$ related domains, we refer to $D_k$ as a dataset with $N_k$ samples for domain $k$. $D_k = \{(S_i^{(k)},Y_i^{(k)})\}_{i=1}^{N_k}$,
% \begin{equation}
% ,
% \end{equation}
where $S_i^{(k)}$ and $Y_i^{(k)}$ represent a sequence of sentences and the corresponding label sequence from a document of domain $k$, respectively.
The goal is to estimate the conditional probability $P(Y|S)$ by utilizing the complementarities among different domains.

% \vspace{-5pt}

% TBC:模型部分可以先写上啦，，，
\paragraph{Model$^{I}_{Base}$}
% Base Model: aggregate five indomain dataset into a training set
This is a simple but effective model for multi-domain learning, in which all domains are aggregated together and will be further used for training a set of shared parameters.
Notably, domains in this model are not explicitly informed of their differences.

Therefore, the loss function of each domain can be written as:
\begin{align}
    \mathcal{L}^{(k)}_{I} = {L}(\textsc{Basic}(\mathbf{S}^{(k)},\theta^{(s)}),\mathbf{Y}^{(k)})
    \label{eqn:model-1}
\end{align}
where \textsc{Basic} denotes our CNN-Transformer encoder framework (As described in Section \ref{sec:basicmodel}). $\theta^{(s)}$ means that all domains share the same parameters.

\textbf{Analysis:} The above model benefits from the \textit{joint training} strategy, which can allow a monolithic model to learn shared features from different domains.
However, it is not sufficient to alleviate the domain shift problem, because two potential limitations remain:
1) The joint model is not aware of the differences across domains, which would lead to poor performance on in-task evaluation since some task-specific features shared by other tasks.
2) Negative transferring might happened on new domains.
Next, we will study three different approaches to address the above problems.

% \vspace{-5pt}

% introduce BERT
\paragraph{Model$^{II}_{BERT}$} %Domain-aware by BERT
% Model4: Semi-supervised Pre-training

More recently, unsupervised pre-training has achieved massive success in NLP community \cite{devlin2018bert,peters2018deep}, which usually provides tremendous external knowledge.
However, there are few works on building the connection between large-scale pre-trained models and multi-domain learning.
In this model, we explore how the external knowledge unsupervised pre-trained models bring can contribute to multi-domain learning and new domain adaption \footnote{Concurrent with our work, \citet{Radford2018} also apply pre-trained language model to a wide range of NLP tasks in a zero-shot setting. We will discuss the differences in the related work section. }.

We achieve this by pre-training our basic model $Model^{I}_{Base}$ with BERT \cite{devlin2018bert}, which is one of the most successful learning frameworks.  Then we investigate if BERT can provide domain information and bring the model good domain adaptability. To avoid introducing new structures, we use the feature-based BERT with its parameters fixed.

\textbf{Analysis:}
This model instructs the processing of multi-domain learning by utilizing external pre-trained knowledge. Another perspective is to address this problem algorithmically.

% \vspace{-5pt}
\paragraph{Model$^{III}_{Tag}$}
%With above analysis in mind, we next we propose two improved domain-aware models.
The domain type can also be introduced directly as a feature vector, which can augment learned representations with domain-aware ability.

Specifically, each domain tag  $C^{(k)}$ will be embedded into a low dimensional real-valued vector and then be concatenated with sentence embedding $\mathbf{s^{(k)}_i}$. The loss function can be formulated as:
% and fed into the document encoder to get a contextualized sentence representation.
\begin{align}
    \mathcal{L}^{(k)}_{III} = {L}(\textsc{Basic}(\mathbf{S}^{(k)},C^{(k)},\theta^{(s)}),\mathbf{Y}^{(k)})
    \label{eqn:model-3}
\end{align}

It is worth noting that, on unseen domains, the information of real domain tags is not available. Thus we design a domain tag `$\mathfrak{X}$' for unknown domains and randomly relabeled examples with it during training.
Since the real tag of the data tagged with `$\mathfrak{X}$' may be any source domain, this embedding will force the model to learn the shared features and makes it more adaptive to unseen domains. In the experiment, this improves the performance on both source domains and target domains.
%
% and reassign 10\% training examples to it randomly.

\textbf{Analysis:} This domain-aware model makes it possible to learn domain-specific features, while it still suffers from the negative transfer problem since private and shared features are entangled in shared space \citep{bousmalis2016domain,liu2017adversarial}. Specifically, each domain has permission to modify shared parameters, which makes it easier to update parameters along different directions.

\begin{figure}
  \centerline{
    \includegraphics[width=1\linewidth]{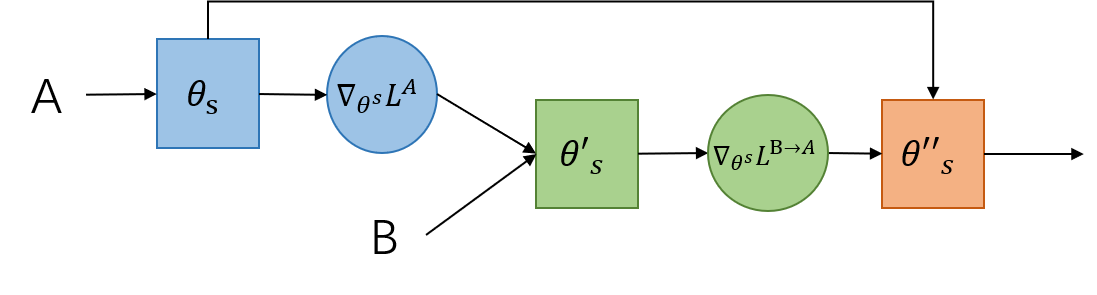}}
  \caption{The gradient update mechanism of the meta learning strategy of Model$^{IV}_{Meta}$. }
  \label{fig:model-4}
\end{figure}
\setlength{\belowcaptionskip}{-0.7cm}

% \vspace{-2pt}
\paragraph{Model$^{IV}_{Meta}$}
% Model 3: Meta-Learning
In order to overcome the above limitations, we try to bridge the communication gap between different domains when updating shared parameters via meta-learning \cite{finn2017model,li2017learning,liu2018meta}.

Here, the introduced communicating protocol claims that each domain should tell others what its updating details (gradients) are. Through its different updating behaviors of different domains can be more consistent.

Formally, given a main domain $A$ and an auxiliary domain $B$, the model will first compute the gradients of A $\nabla_{\theta} \mathcal{L}^{A}$ with regard to the model parameters $\theta$. Then the model will be updated with the gradients and calculate the gradients of B.

Our objective is to produce maximal performance on sample $(S^{(B)},Y^{(B)})$:
\begin{equation}
    \mathcal{L}^{A\leftarrow B}  =   \min_{\theta} L(S^{(B)},Y^{(B)}, \nabla_{\theta} \mathcal{L}^{A})
%\vspace{-0.05cm}
\end{equation}

So, the loss function for each domain can be finally written as:
\begin{align}
    \mathcal{L}^{(k)}_{IV}  = \gamma \mathcal{L}^{(k)} + (1 - \gamma) \sum_{j \neq k} \mathcal{L}^{k\leftarrow j}
    \label{eqn:model-4}
\end{align}
where $\gamma$ $(0 \leq \gamma \leq 1)$ is the weight coefficient and $\mathcal{L}$ can be instantiated as $\mathcal{L}_{I}$ (Eqn. \ref{eqn:model-1}), $\mathcal{L}_{II}$ or $\mathcal{L}_{III}$ (Eqn. \ref{eqn:model-3}).
% Figure \ref{fig:model-4} illustrates the updating mechanism.

\textbf{Analysis}: To address the multi-domain learning task and the adaptation to new domains,
Model$^{II}_{BERT}$, Model$^{III}_{Tag}$, Model$^{IV}_{Meta}$ take different angles. Specifically, Model$^{II}_{BERT}$ utilizes a large-scale pre-trained model while Model$^{III}_{Tag}$ proposes to introduce domain type information explicitly.
Lastly,  Model$^{IV}_{Meta}$ is designed to update parameters more consistently, by adjusting the gradient direction of the main domain A with the auxiliary domain B during training. This mechanism indeed purifies the shared feature space via filtering out the domain-specific features which only benefit A.

% TBC: 实验各种细节操作 也可以写上啦
\section{Experiment}
\label{sec:exper}

We investigate the effectiveness of the above four strategies under three evaluation settings: \textsc{in-domain}, \textsc{out-of-domain} and \textsc{cross-dataset}. These settings make it possible to explicitly evaluate models both on the quality of domain-aware text representation and on their adaptation ability to derive reasonable representations in unfamiliar domains.

\subsection{Experiment Setup}
\label{sec:impl}
We perform our experiments mainly on our multi-domain \textsc{MULTI-SUM} dataset. Source domains are defined as the first five domains (\textsc{in-domain}) in Table \ref{tab:twosubsets} and the other domains (\textsc{out-of-domain}) are totally invisible during training. The evaluation under the \textsc{in-domain} setting tests the model ability to learn different domain distribution on a multi-domain set and later \textsc{out-of-domain} investigates how models perform on unseen domains. We further make use of CNN/DailyMail as a \textsc{cross-dataset} evaluation environment to provide a larger distribution gap.

We use Model$^{I}_{Basic}$ as a baseline model, build Model$^{II}_{BERT}$ with feature-based BERT and Model$^{III}_{Tag}$ with domain embedding on it. We further develop Model$^{III}_{Tag}$ as the instantiation of Model$^{IV}_{Meta}$. For the detailed dataset statistics, model settings and hyper-parameters, the reader can refer to Appendix.

\begin{table}[!t]\footnotesize\setlength{\tabcolsep}{1pt}
% \begin{tabular}{l*{14}{c}}
\centering
\begin{tabular}{lcccc}
\toprule
\textbf{Domains} &
{Model$^{I}_{Basic}$} &
{Model$^{II}_{BERT}$} &
{Model$^{III}_{Tag}$} &
{Model$^{IV}_{Meta}$} \\
\midrule
\multicolumn{3}{l}{\textsc{In-domain Setting}} \\
\midrule
FN    & 49.13  & 49.70 & 49.54  & 49.06 \\
CNN   & 41.84  & 41.59  & 42.14  & 41.73  \\
MA    & 35.40  & 36.46  & 35.51  & 34.85  \\
NYT   & 30.68  & 31.92  & 30.70  & 30.20  \\
WTP   & 23.88  & 24.47  & 24.02  & 23.51  \\
\midrule
\rowcolor[rgb]{ .851,  .851,  .851} $Average$ & 34.24 & \textbf{35.04}  & 34.40 &  33.95 \\
\midrule
\midrule
\multicolumn{3}{l}{\textsc{Out-of-domain Setting}} \\
\midrule
NTDN  & 48.62  & 49.37  & 49.44  & 49.47  \\
WSJ   & 38.55  & 39.63  & 39.00  & 38.93  \\
USAT  & 28.85  & 29.57 & 29.09  & 28.99 \\
TG    & 24.15  & 24.89  & 24.10  & 24.18 \\
TIME  & 21.67  & 22.45  & 22.20  & 22.12  \\
\midrule
\rowcolor[rgb]{ .851,  .851,  .851} $Average$ & 32.78  & \textbf{33.68} & 33.17 & 33.24 \\
\midrule
$\Delta R$ & $1.47\downarrow$  & $1.35\downarrow$    & $1.24\downarrow$    & \textcolor[rgb]{ 1,  0,  0}{\textbf{$0.71\downarrow$}}  \\
\midrule
\midrule
\multicolumn{3}{l}{\textsc{Cross-dataset Setting}} \\
\midrule
\rowcolor[rgb]{ .851,  .851,  .851} CNN/DM & 40.11  & 39.82  & 40.28  & \textbf{40.30}  \\
\bottomrule
%----------------------------------------------------------------------
\end{tabular}
\caption{Rouge-1 performance of our four learning strategies on the \textsc{MULTI-SUM} dataset.  $\Delta R = |Rouge(\textsc{in-domain}) - Rouge(\textsc{out-of-domain})|$ .
A smaller $\Delta R$ indicates corresponding model has a better generalization ability.
Bold numbers are the best results, and red ones indicate the minimum performance gap between source and target domains. The grey rows show the models' average performance under three evaluation settings.}
\label{tab:4models}
\end{table}
\setlength{\belowcaptionskip}{-12pt}
\vspace{-5pt}

\begin{table}[htbp]\small
 \centering
   \begin{tabular}{lccc}
   \toprule
   Model & R-1 & R-2 & R-L \\
   \midrule
   Lead-3 \cite{See2017} & 40.34 & 17.70 & 36.57 \\
   \midrule
   \citet{Narayan2018rank} & 40.00 & 18.20 & 36.60 \\
   \citet{Zhang2018} & 41.05 & 18.77 & 37.54 \\
   \citet{Chen2018fast} & 41.47 & 18.72 & 37.76 \\
   \citet{Dong2018banditSum} & 41.50 & 18.70 & 37.60 \\
   \citet{Zhou2018} & \textbf{41.59} & \textbf{19.01} & \textbf{37.98} \\
   \midrule
   Our basic model & 41.33 & 18.83 & 37.65 \\
   Basic model + Tag & {0.13$\uparrow$} & {0.05$\uparrow$} & {0.10$\uparrow$} \\
   Basic model + Meta & {0.07$\downarrow$} & {0.02$\downarrow$} & {0.05$\downarrow$} \\
   Basic model + BERT & {0.93$\uparrow$} & {0.90$\uparrow$} & {0.97$\uparrow$} \\
   Basic model + BERT + Tag & {0.97$\uparrow$} & {0.93$\uparrow$} & {1.01$\uparrow$} \\
   \bottomrule
   \end{tabular}%
 \caption{Comparison between our strategies with other extractive summarization models on non-anonymized CNN/Daily Mail provided by \citet{See2017}. The red up-arrows indicate performance improvement over our base model, and the green down-arrows denote the degradation.}
 \label{tab:cnndm}%
\end{table}%

\subsection{Quantitative Results}
\label{sec:quanti}
We compare our models by ROUGE-1 scores in Table \ref{tab:4models}. Note that we select two sentences for \textsc{MULTI-SUM} domains and three sentences for CNN/Daily Mail due to the different average lengths of reference summaries.

\vspace{-2pt}

%  Should we train a monolithic model, ignoring the existence of the domain or train domain-ware model for summarization tasks?
% \paragraph{Monolithic vs Domain-aware}
\paragraph{Model$^{I}_{Basic}$ vs Model$^{III}_{Tag}$}
From Table \ref{tab:4models}, we observe that the domain-aware model outperforms the monolithic model under both \textsc{in-domain} and \textsc{out-of-domain} settings. The significant improvement of \textsc{in-domain} demonstrates domain information is effective for summarization models trained on multiple domains. Meanwhile, the superior performance on \textsc{out-of-domain} further illustrates that, the awareness of domain difference also benefits under the zero-shot setting. This might suggest that the domain-aware model could capture domain-specific features by domain tags and have learned domain-invariant features at the same time, which can be transferred to unseen domains.
%The shared features are relatively more domain-invariant and can also work well on unseen domains.

\vspace{-2pt}

\paragraph{Model$^{I}_{Basic}$ vs Model$^{IV}_{Meta}$}
Despite a little drop under \textsc{in-domain} setting, the narrowed performance gap, as shown in $\Delta R$ of Table \ref{tab:4models},
% \ref{fig:deltaR},
% Table \ref{tab:4models},
indicates Model$^{IV}_{Meta}$ has better generalization ability as a compensation.
The performance decline mainly lies in the more consistent way to update parameters, which purifies shared feature space at the expense of filtering out some domain-specific features.
The excellent results under \textsc{cross-dataset} settings further suggest the meta-learning strategy successfully improve the model transferability not only among the domains of \textsc{MULTI-SUM} but also across different datasets.
% This indicates that sharing updating details of different domains is more effective when adapting to new domains.

\begin{figure*}
  \centering
    \includegraphics[width=1\textwidth]{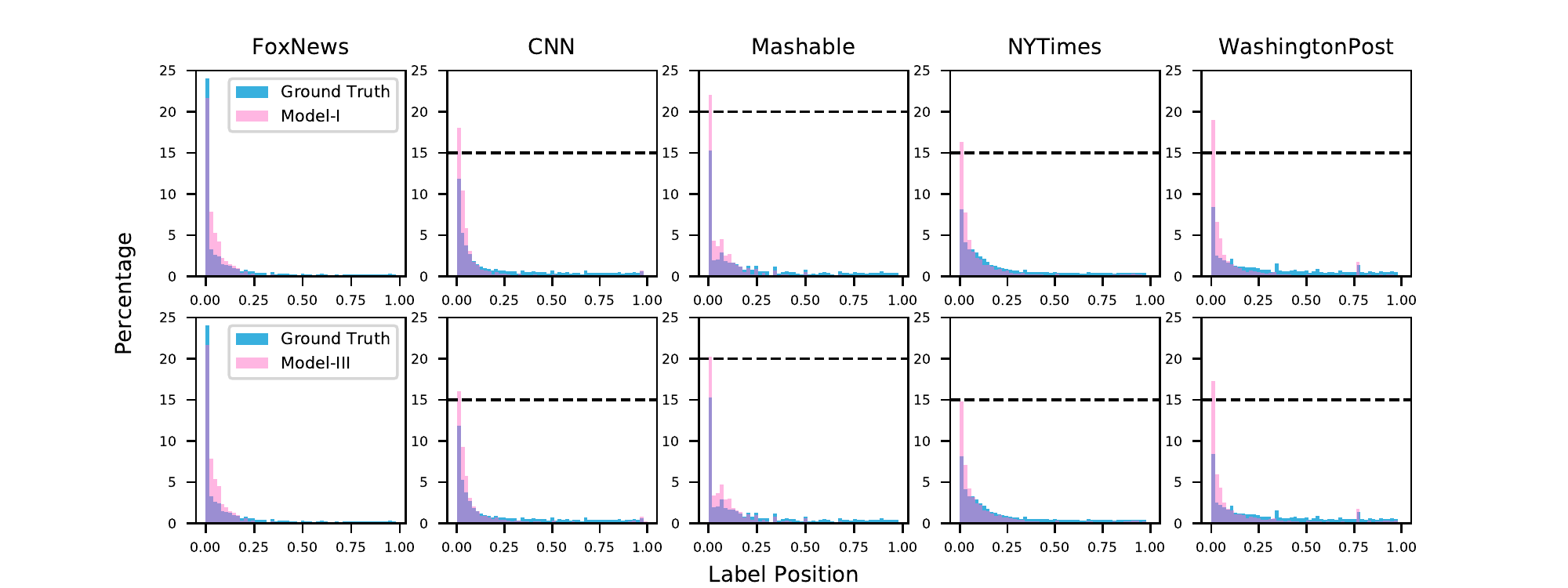}
  \caption{Relative position of selected sentence in the original document across five source domains. We overlap the ground truth labels with the model results in order to highlight the differences. The two rows correspond to Model-I and Model-III in Section \ref{sec:models}.}
  \label{fig:AllModelPos}
\end{figure*}

\vspace{-2pt}

% How to learn an summarization model which can perform better both  under \textsc{in-domain} and \textsc{out-of-domain} settings?
\paragraph{Model$^{II}_ {BERT}$}
% Model$^{II}_ {BERT}$ outperforms all other models under both \textsc{in-domain} and \textsc{out-of-domain} settings, however, its performance on \textsc{cross-dataset} setting is poor.
Supported by the smaller $\Delta R$ compared with Model$^{I}_{Base}$, we can draw the conclusion that BERT shows some domain generalization ability
\footnote{We give a specific experiment and analyze why Model$^{II}_ {BERT}$ with BERT can achieve domain generalization in Appendix.}
within \textsc{MULTI-SUM}. However, this ability is inferior to Model$^{III}_{Tag}$ and Model$^{IV}_{Meta}$, which further leads to the worse performance on \textsc{cross-dataset}. Thus we cannot attribute its success in \textsc{MULTI-SUM} to the ability to address multi-domain learning nor domain adaptation. Instead, we suppose the vast external knowledge of BERT provides its superior ability for feature extraction. That causes Model$^{II}_ {BERT}$ to overfit \textsc{MULTI-SUM} and perform excellently across all domains, but fails on the more different dataset CNN/Daily Mail.

This observation also suggests that although unsupervised pre-trained models are powerful enough \cite{Radford2018}, still, it can not take place the role of supervised learning methods (i.e. Model$^{III}_{Tag}$ and Model$^{IV}_{Meta}$), which is designed specifically for addressing multi-domain learning and new domain adaptation.

%Notably, from the following analysis, we know $\Delta R$ can measure the ability to generalize to unfamiliar position distribution. However

% \vspace{-5pt}
\paragraph{Analysis of Different Model Choices }
\label{sec:model_analysis}
To summarize,
Model$^{III}_ {Tag} $ is a simple and efficient method, which can achieve good performance under in-domain setting and shows certain generalization ability on the unseen domain.
Model$^{IV}_ {Meta} $ shows the best generalization ability at the cost of relatively lower in-domain performance. Therefore, using Model$^{IV}_ {Meta} $ is not a good choice if in-domain performance matters for end users.
Model$^{II}_ {BERT} $ can achieve the best performance under in-domain settings at expense of training time and shows worse generalization ability than Model$^{IV}_ {Meta} $.
If the training time is not an issue, Model$^{II}_ {BERT} $ could be a good supplement for other methods.

\subsection{Results on CNN/DailyMail}
Inspired by such observations, we further employ our four learning strategies to the mainstream summarization dataset CNN/DailyMail \cite{See2017}, which also includes two different data sources: \textit{CNN} and \textit{DailyMail}. We use the publication as the domain and train our models on its 28w training set. As Table \ref{tab:cnndm} shows, our basic model has comparable performance with other extractive summarization models. Besides, the publication tags can improve ROUGE scores significantly by 0.13 points in ROUGE-1 and the meta learning strategy does not show many advantages when dealing with in-domain examples, what we have expected. BERT with tags achieves the best performance, although the performance increment is not as much as what publication tags bring to the basic model, which we suppose that BERT itself has contained some degree of domain information.

\subsection{Qualitative Analysis}
\label{sec:qual}
% TBC: 这部分是最最重要的，就是我们还要设计一些实验，可以定量也可以定性，甚至是可视化，目的就是对我们的模型或关注问题有更细致的探究；这个从现在开始就可以好好想想啦；
We furthermore design several experiments to probe into some potential factors that might contribute to the superior performance of domain-aware models over the monolithic basic model.

\vspace{-5pt}
\paragraph{Label Position} Sentence position is a well known and powerful feature, especially for extractive summarization \cite{Kedzie2018} \footnote{We plot the density histogram of the relative locations of ground truth labels for both source and target domains and attach it in Appendix. Compared with Table \ref{tab:results_cross_domains}, we can find that the relative position of ground truth labels is closely related to ROUGE performance of the basic model.}.
We compare the relative position of sentences selected by our models with the ground truth labels on source domains to investigate how well these models fit the distribution and whether they can distinguish between domains. We select the most representative models Model$^{I}_{Base}$ and Model$^{III}_{Tag}$ illustrated in Figure \ref{fig:AllModelPos} \footnote{The whole picture in the Appendix illustrates the four models performance.}.
% To be specific, domains with labels concentrated on the front parts of documents have a more strong lead bias \cite{Kim2018abs} and thus are easier to learn. At the same time, uniform distributions imply more dispersed labels and the basic model trained on these domains usually show poor performance.

% It is reasonable to conjecture that the label relative position contributes to the domain shift encountered in text summarization.

The percentage of the first sentence on \textit{FoxNews} is significantly higher than others:
(1) Unaware of different domains, Model$^{I}_{Base}$ learns a similar distribution for all domains and is seriously affected by this extreme distribution. In its density histogram, the probability of the first sentence being selected is much higher than the ground truth on the other four domains.
(2) Compared with Model$^{I}_{Base}$, domain-aware models are more robust by learning different relative distributions for different domains. Model$^{III}_{Tag}$ constrains the extreme trend especially obviously on \textit{CNN} and \textit{Mashable}.

% Model-III and Model-IV both constrain the extreme lead bias.
% (3) Model-II is more consistent with the distribution of \textit{Mashable} between 0.0 and 0.1 and is the only one that learn the distinctive bias of \textit{WashingtonPost} around 0.1, thus it achieves best results on the \textsc{in-domain} settings. However, it is still seriously affected as Model-I.

\begin{figure}[t]
  \centering
  \subfigure[In-domain]{
    \label{fig:test1}
    \includegraphics[width=0.30\linewidth]{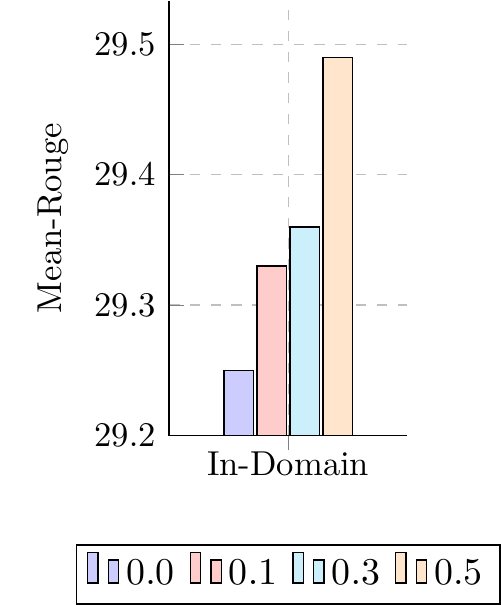}
  }
  \subfigure[Out-of-domain]{
    \label{fig:test2}
    \includegraphics[width=0.30\linewidth]{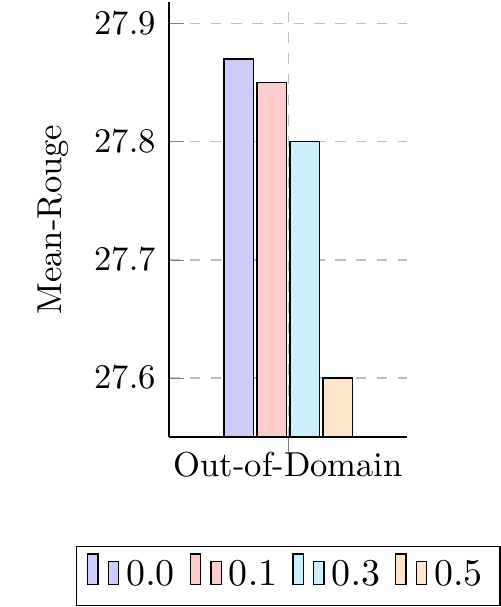}
  }
  \subfigure[Cross-Dataset]{
    \label{fig:test3}
    \includegraphics[width=0.30\linewidth]{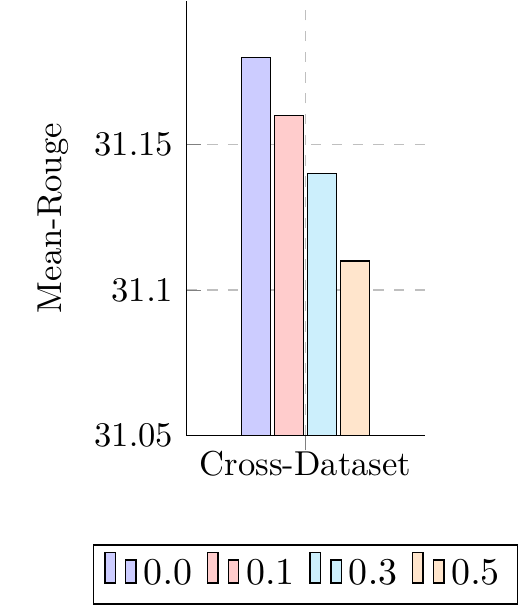}
  }
 \caption{Loss weight coefficients $\gamma$ for Model-IV. The y-axis is the mean score of ROUGE-1, ROUGE-2 and ROUGE-L and different bins correspond to different $\gamma$ values.}
 \label{fig:gammas}
\end{figure}
\setlength{\belowcaptionskip}{-2cm}

\vspace{-5pt}
\paragraph{Weight $\gamma$ for Model$^{IV}_{Meta}$}

We investigate several $\gamma$ to further probe into the performance of Model$^{IV}_{Meta}$. In Eqn. \ref{eqn:model-4}, $\gamma$ is the weight coefficient of main domain A. When $\gamma=0$, the model ignores A and focuses on the auxiliary domain B and when $\gamma=1$ it is trained only on the loss of main domain A (the same as the instantiation Model$^{III}_{Tag}$). As Figure \ref{fig:gammas} shows, with the increase of $\gamma$, the Rouge scores rise on \textsc{in-domain} while decline on \textsc{out-of-domain} and \textsc{cross-dataset}. The performances under \textsc{in-domain} settings prove that the import of the auxiliary domain hurts the model ability to learn domain-specific features. However, results under both \textsc{out-of-domain} and \textsc{cross-dataset} settings indicate the loss of B, which is informed of A's gradient information, helps the model to learn more general features, thus improving the generalization ability.

\section{Related Work}
%Our work touches on the following strands of research:
We briefly outline connections and differences to the following related lines of research.
%Due to space constraints, we cannot do justice to the complete literature

\vspace{-5pt}
\paragraph{Domains in Summarization} There have been several works in summarization exploring the concepts of domains. \citet{cheung2013towards} explored domain-specific knowledge and associated it as template information. \citet{hua2017pilot} investigated domain adaptation in abstractive summarization and found the content selection is transferable to a new domain. \citet{Gehrmann2018bottom} trained a selection mask for abstractive summarization and proved it has excellent adaptability. However, previous works just investigated models trained on a single domain and did not explore multi-domain learning in summarization.

\vspace{-5pt}
\paragraph{Multi-domain Learning (MDL) \& Domain Adaptation (DA)}
% This paper shares the properties of both multi-domain learning and domain adaptation.
We focus on the testbed that requires both training and evaluating performance on a set of domains.
Therefore, we care about two questions:
1) how to learn a model when the training set contains multiple domains -- involving MDL.
2) how to adapt the multi-domain model to new domains -- involving DA.
Beyond the investigation of some effective approaches like existing works, we have first verified how domain shift influences the summarization tasks.

\vspace{-5pt}
\paragraph{Semi-supervised Pre-training for Zero-shot Transfer}
%Supervised or unsupervised pre-trained models can be used to fine-tune down-stream tasks.
It has a long history of fine-tuning downstream tasks with supervised or unsupervised pre-trained models \cite{le2014distributed,devlin2018bert,peters2018deep}.
However, there is a rising interest in applying large-scale pre-trained models to zero-shot transfer learning \cite{Radford2018}.
% Concurrent with our work, \cite{Radford2018} first apply pre-trained language model to a large body of NLP tasks.
Different from the above works, we focus on addressing domain shift and generalization problem. One of our explored methods is semi-supervised pre-training, which combines supervised and unsupervised approaches to achieve zero-shot transfer.

\section{Conclusion}
In this paper, we explore publication in the context of the domain and investigate the domain shift problem in summarization. When verified its existence, we propose to build a multi-domain testbed for summarization that requires both training and measuring performance on a set of domains. Under these new settings, we propose four learning schemes to give a preliminary explore in characteristics of different learning strategies when dealing with multi-domain summarization tasks.

\section*{Acknowledgment}
We thank Jackie Chi Kit Cheung for useful comments and discussions.
The research work is supported by National Natural Science Foundation of China (No. 61751201 and 61672162),
Shanghai Municipal Science and Technology Commission (16JC1420401 and 17JC1404100),
Shanghai Municipal Science and Technology Major Project(No.2018SHZDZX01)and ZJLab.

\bibliography{./nlp}

\begin{thebibliography}{43}
\expandafter\ifx\csname natexlab\endcsname\relax\def\natexlab#1{#1}\fi

\bibitem[{Blei et~al.(2003)Blei, Ng, and Jordan}]{blei2003latent}
David~M Blei, Andrew~Y Ng, and Michael~I Jordan. 2003.
\newblock Latent dirichlet allocation.
\newblock \emph{Journal of machine Learning research}, 3(Jan):993--1022.

\bibitem[{Blitzer et~al.(2007)Blitzer, Dredze, and
  Pereira}]{blitzer2007biographies}
John Blitzer, Mark Dredze, and Fernando Pereira. 2007.
\newblock Biographies, bollywood, boom-boxes and blenders: Domain adaptation
  for sentiment classification.
\newblock In \emph{Proceedings of the 45th annual meeting of the association of
  computational linguistics}, pages 440--447.

\bibitem[{Bousmalis et~al.(2016)Bousmalis, Trigeorgis, Silberman, Krishnan, and
  Erhan}]{bousmalis2016domain}
Konstantinos Bousmalis, George Trigeorgis, Nathan Silberman, Dilip Krishnan,
  and Dumitru Erhan. 2016.
\newblock Domain separation networks.
\newblock In \emph{Advances in Neural Information Processing Systems}, pages
  343--351.

\bibitem[{Cao et~al.(2017)Cao, Li, Li, and Wei}]{cao2017improving}
Ziqiang Cao, Wenjie Li, Sujian Li, and Furu Wei. 2017.
\newblock \href {http://arxiv.org/abs/1611.09238} {{Improving Multi-Document
  Summarization via Text Classification}}.
\newblock \emph{Proceedings of the 31th Conference on Artificial Intelligence
  (AAAI 2017)}.

\bibitem[{Chen and Bansal(2018)}]{Chen2018fast}
Yen-Chun Chen and Mohit Bansal. 2018.
\newblock Fast abstractive summarization with reinforce-selected sentence
  rewriting.
\newblock In \emph{Proceedings of the 56th Annual Meeting of the Association
  for Computational Linguistics (Volume 1: Long Papers)}, volume~1, pages
  675--686.

\bibitem[{Cheng and Lapata(2016)}]{cheng2016neural}
Jianpeng Cheng and Mirella Lapata. 2016.
\newblock Neural summarization by extracting sentences and words.
\newblock In \emph{Proceedings of the 54th Annual Meeting of the Association
  for Computational Linguistics (Volume 1: Long Papers)}, volume~1, pages
  484--494.

\bibitem[{Cheung and Penn(2013{\natexlab{a}})}]{cheung2013probabilistic}
Jackie Chi~Kit Cheung and Gerald Penn. 2013{\natexlab{a}}.
\newblock Probabilistic domain modelling with contextualized distributional
  semantic vectors.
\newblock In \emph{Proceedings of the 51st Annual Meeting of the Association
  for Computational Linguistics (Volume 1: Long Papers)}, volume~1, pages
  392--401.

\bibitem[{Cheung and Penn(2013{\natexlab{b}})}]{cheung2013towards}
Jackie Chi~Kit Cheung and Gerald Penn. 2013{\natexlab{b}}.
\newblock Towards robust abstractive multi-document summarization: A caseframe
  analysis of centrality and domain.
\newblock In \emph{Proceedings of the 51st Annual Meeting of the Association
  for Computational Linguistics (Volume 1: Long Papers)}, volume~1, pages
  1233--1242.

\bibitem[{Devlin et~al.(2018)Devlin, Chang, Lee, and
  Toutanova}]{devlin2018bert}
Jacob Devlin, Ming-Wei Chang, Kenton Lee, and Kristina Toutanova. 2018.
\newblock Bert: Pre-training of deep bidirectional transformers for language
  understanding.
\newblock \emph{arXiv preprint arXiv:1810.04805}.

\bibitem[{Dong et~al.(2018)Dong, Shen, Crawford, van Hoof, and
  Cheung}]{Dong2018banditSum}
Yue Dong, Yikang Shen, Eric Crawford, Herke van Hoof, and Jackie Chi~Kit
  Cheung. 2018.
\newblock \href {http://arxiv.org/abs/1809.09672} {{BanditSum: Extractive
  Summarization as a Contextual Bandit}}.
\newblock In \emph{Empirical Methods in Natural Language Processing (EMNLP)}.

\bibitem[{Finn et~al.(2017)Finn, Abbeel, and Levine}]{finn2017model}
Chelsea Finn, Pieter Abbeel, and Sergey Levine. 2017.
\newblock Model-agnostic meta-learning for fast adaptation of deep networks.
\newblock In \emph{International Conference on Machine Learning}, pages
  1126--1135.

\bibitem[{Gehrmann et~al.(2018)Gehrmann, Deng, and Rush}]{Gehrmann2018bottom}
Sebastian Gehrmann, Yuntian Deng, and Alexander~M. Rush. 2018.
\newblock \href {http://arxiv.org/abs/1808.10792} {{Bottom-Up Abstractive
  Summarization}}.
\newblock In \emph{Empirical Methods in Natural Language Processing (EMNLP)}.

\bibitem[{Gopalan et~al.(2011)Gopalan, Li, and Chellappa}]{gopalan2011domain}
Raghuraman Gopalan, Ruonan Li, and Rama Chellappa. 2011.
\newblock Domain adaptation for object recognition: An unsupervised approach.
\newblock In \emph{2011 international conference on computer vision}, pages
  999--1006. IEEE.

\bibitem[{Grusky et~al.(2018)Grusky, Naaman, and Artzi}]{grusky2018newsroom}
Max Grusky, Mor Naaman, and Yoav Artzi. 2018.
\newblock Newsroom: A dataset of 1.3 million summaries with diverse extractive
  strategies.
\newblock In \emph{Proceedings of the 2018 Conference of the North American
  Chapter of the Association for Computational Linguistics: Human Language
  Technologies, Volume 1 (Long Papers)}, volume~1, pages 708--719.

\bibitem[{Haghighi and Vanderwende(2009)}]{haghighi2009exploring}
Aria Haghighi and Lucy Vanderwende. 2009.
\newblock Exploring content models for multi-document summarization.
\newblock In \emph{Proceedings of Human Language Technologies: The 2009 Annual
  Conference of the North American Chapter of the Association for Computational
  Linguistics}, pages 362--370. Association for Computational Linguistics.

\bibitem[{Hermann et~al.(2015)Hermann, Kocisky, Grefenstette, Espeholt, Kay,
  Suleyman, and Blunsom}]{hermann2015teaching}
Karl~Moritz Hermann, Tomas Kocisky, Edward Grefenstette, Lasse Espeholt, Will
  Kay, Mustafa Suleyman, and Phil Blunsom. 2015.
\newblock Teaching machines to read and comprehend.
\newblock In \emph{Advances in Neural Information Processing Systems}, pages
  1693--1701.

\bibitem[{Hua and Wang(2017)}]{hua2017pilot}
Xinyu Hua and Lu~Wang. 2017.
\newblock A pilot study of domain adaptation effect for neural abstractive
  summarization.
\newblock \emph{arXiv preprint arXiv:1707.07062}.

\bibitem[{Isonuma et~al.(2017)Isonuma, Fujino, Mori, Matsuo, and
  Sakata}]{isonuma2017extractive}
Masaru Isonuma, Toru Fujino, Junichiro Mori, Yutaka Matsuo, and Ichiro Sakata.
  2017.
\newblock Extractive summarization using multi-task learning with document
  classification.
\newblock In \emph{Proceedings of the 2017 Conference on Empirical Methods in
  Natural Language Processing}, pages 2101--2110.

\bibitem[{Joshi et~al.(2012)Joshi, Cohen, Dredze, and
  Ros{\'e}}]{joshi2012multi}
Mahesh Joshi, William~W Cohen, Mark Dredze, and Carolyn~P Ros{\'e}. 2012.
\newblock Multi-domain learning: when do domains matter?
\newblock In \emph{Proceedings of the 2012 Joint Conference on Empirical
  Methods in Natural Language Processing and Computational Natural Language
  Learning}, pages 1302--1312. Association for Computational Linguistics.

\bibitem[{Kedzie et~al.(2018)Kedzie, Mckeown, and Daum}]{Kedzie2018}
Chris Kedzie, Kathleen Mckeown, and Hal Daum. 2018.
\newblock {Content Selection in Deep Learning Models of Summarization}.
\newblock In \emph{Empirical Methods in Natural Language Processing (EMNLP)}.

\bibitem[{Le and Mikolov(2014)}]{le2014distributed}
Quoc~V. Le and Tomas Mikolov. 2014.
\newblock Distributed representations of sentences and documents.
\newblock In \emph{Proceedings of ICML}.

\bibitem[{Li et~al.(2017)Li, Yang, Song, and Hospedales}]{li2017learning}
Da~Li, Yongxin Yang, Yi-Zhe Song, and Timothy~M Hospedales. 2017.
\newblock Learning to generalize: Meta-learning for domain generalization.
\newblock \emph{arXiv preprint arXiv:1710.03463}.

\bibitem[{Li and Zong(2008)}]{li2008multi}
Shoushan Li and Chengqing Zong. 2008.
\newblock Multi-domain sentiment classification.
\newblock In \emph{Proceedings of the 46th Annual Meeting of the Association
  for Computational Linguistics on Human Language Technologies: Short Papers},
  pages 257--260. Association for Computational Linguistics.

\bibitem[{Lin and Hovy(2003)}]{lin2003automatic}
Chin-Yew Lin and Eduard Hovy. 2003.
\newblock Automatic evaluation of summaries using n-gram co-occurrence
  statistics.
\newblock In \emph{Proceedings of the 2003 Human Language Technology Conference
  of the North American Chapter of the Association for Computational
  Linguistics}.

\bibitem[{Liu and Huang(2018)}]{liu2018meta}
Pengfei Liu and Xuanjing Huang. 2018.
\newblock Meta-learning multi-task communication.
\newblock \emph{arXiv preprint arXiv:1810.09988}.

\bibitem[{Liu et~al.(2017)Liu, Qiu, and Huang}]{liu2017adversarial}
Pengfei Liu, Xipeng Qiu, and Xuanjing Huang. 2017.
\newblock Adversarial multi-task learning for text classification.
\newblock In \emph{Proceedings of the 55th Annual Meeting of the Association
  for Computational Linguistics (Volume 1: Long Papers)}, volume~1, pages
  1--10.

\bibitem[{Nallapati et~al.(2017)Nallapati, Zhai, and
  Zhou}]{nallapati2017summarunner}
Ramesh Nallapati, Feifei Zhai, and Bowen Zhou. 2017.
\newblock Summarunner: A recurrent neural network based sequence model for
  extractive summarization of documents.
\newblock In \emph{Thirty-First AAAI Conference on Artificial Intelligence}.

\bibitem[{Napoles et~al.(2012)Napoles, Gormley, and
  Van~Durme}]{napoles2012annotated}
Courtney Napoles, Matthew Gormley, and Benjamin Van~Durme. 2012.
\newblock Annotated gigaword.
\newblock In \emph{Proceedings of the Joint Workshop on Automatic Knowledge
  Base Construction and Web-scale Knowledge Extraction}, pages 95--100.
  Association for Computational Linguistics.

\bibitem[{Narayan et~al.(2018{\natexlab{a}})Narayan, Cohen, and
  Lapata}]{narayan2018don}
Shashi Narayan, Shay~B Cohen, and Mirella Lapata. 2018{\natexlab{a}}.
\newblock Don't give me the details, just the summary! topic-aware
  convolutional neural networks for extreme summarization.
\newblock \emph{arXiv preprint arXiv:1808.08745}.

\bibitem[{Narayan et~al.(2018{\natexlab{b}})Narayan, Cohen, and
  Lapata}]{Narayan2018rank}
Shashi Narayan, Shay~B. Cohen, and Mirella Lapata. 2018{\natexlab{b}}.
\newblock \href {http://arxiv.org/abs/1802.08636} {{Ranking Sentences for
  Extractive Summarization with Reinforcement Learning}}.

\bibitem[{Paulus et~al.(2017)Paulus, Xiong, and Socher}]{paulus2017deep}
Romain Paulus, Caiming Xiong, and Richard Socher. 2017.
\newblock A deep reinforced model for abstractive summarization.
\newblock \emph{arXiv preprint arXiv:1705.04304}.

\bibitem[{Peters et~al.(2018)Peters, Neumann, Iyyer, Gardner, Clark, Lee, and
  Zettlemoyer}]{peters2018deep}
Matthew Peters, Mark Neumann, Mohit Iyyer, Matt Gardner, Christopher Clark,
  Kenton Lee, and Luke Zettlemoyer. 2018.
\newblock Deep contextualized word representations.
\newblock In \emph{Proceedings of the 2018 Conference of the North American
  Chapter of the Association for Computational Linguistics: Human Language
  Technologies, Volume 1 (Long Papers)}, volume~1, pages 2227--2237.

\bibitem[{Radford et~al.(2019)Radford, Wu, Child, Luan, Amodei, and
  Sutskever}]{Radford2018}
Alec Radford, Jeffrey Wu, Rewon Child, David Luan, Dario Amodei, and Ilya
  Sutskever. 2019.
\newblock {Language Models are Unsupervised Multitask Learners}.

\bibitem[{Saenko et~al.(2010)Saenko, Kulis, Fritz, and
  Darrell}]{saenko2010adapting}
Kate Saenko, Brian Kulis, Mario Fritz, and Trevor Darrell. 2010.
\newblock Adapting visual category models to new domains.
\newblock In \emph{European conference on computer vision}, pages 213--226.
  Springer.

\bibitem[{Sandhaus(2008)}]{sandhaus2008new}
Evan Sandhaus. 2008.
\newblock The new york times annotated corpus.
\newblock \emph{Linguistic Data Consortium, Philadelphia}, 6(12):e26752.

\bibitem[{See et~al.(2017)See, Liu, and Manning}]{See2017}
Abigail See, Peter~J Liu, and Christopher~D Manning. 2017.
\newblock Get to the point: Summarization with pointer-generator networks.
\newblock In \emph{Proceedings of the 55th Annual Meeting of the Association
  for Computational Linguistics (Volume 1: Long Papers)}, volume~1, pages
  1073--1083.

\bibitem[{Torralba et~al.(2011)Torralba, Efros et~al.}]{torralba2011unbiased}
Antonio Torralba, Alexei~A Efros, et~al. 2011.
\newblock Unbiased look at dataset bias.
\newblock In \emph{CVPR}, volume~1, page~7. Citeseer.

\bibitem[{Vaswani et~al.(2017)Vaswani, Shazeer, Parmar, Uszkoreit, Jones,
  Gomez, Kaiser, and Polosukhin}]{Vaswani2017}
Ashish Vaswani, Noam Shazeer, Niki Parmar, Jakob Uszkoreit, Llion Jones,
  Aidan~N Gomez, {\L}ukasz Kaiser, and Illia Polosukhin. 2017.
\newblock Attention is all you need.
\newblock In \emph{Advances in Neural Information Processing Systems}, pages
  5998--6008.

\bibitem[{Wang et~al.(2018)Wang, Yao, Tao, Zhong, Liu, and
  Du}]{wang2018reinforced}
Li~Wang, Junlin Yao, Yunzhe Tao, Li~Zhong, Wei Liu, and Qiang Du. 2018.
\newblock A reinforced topic-aware convolutional sequence-to-sequence model for
  abstractive text summarization.
\newblock \emph{arXiv preprint arXiv:1805.03616}.

\bibitem[{Wu and Hu(2018)}]{wu2018learning}
Yuxiang Wu and Baotian Hu. 2018.
\newblock Learning to extract coherent summary via deep reinforcement learning.
\newblock In \emph{Thirty-Second AAAI Conference on Artificial Intelligence}.

\bibitem[{Zhang et~al.(2018)Zhang, Lapata, Wei, and Zhou}]{Zhang2018}
Xingxing Zhang, Mirella Lapata, Furu Wei, and Ming Zhou. 2018.
\newblock \href {http://arxiv.org/abs/1808.07187} {{Neural Latent Extractive
  Document Summarization}}.

\bibitem[{Zhong et~al.(2019)Zhong, Liu, Wang, Qiu, and
  Huang}]{zhong2019searching}
Ming Zhong, Pengfei Liu, Danqing Wang, Xipeng Qiu, and Xuan-Jing Huang. 2019.
\newblock Searching for effective neural extractive summarization: What works
  and what’s next.
\newblock In \emph{Proceedings of the 57th Conference of the Association for
  Computational Linguistics}, pages 1049--1058.

\bibitem[{Zhou et~al.(2018)Zhou, Yang, Wei, Huang, Zhou, and Zhao}]{Zhou2018}
Qingyu Zhou, Nan Yang, Furu Wei, Shaohan Huang, Ming Zhou, and Tiejun Zhao.
  2018.
\newblock Neural document summarization by jointly learning to score and select
  sentences.
\newblock In \emph{Proceedings of the 56th Annual Meeting of the Association
  for Computational Linguistics (Volume 1: Long Papers)}, volume~1, pages
  654--663.

\end{thebibliography}
\bibliographystyle{acl_natbib}

\end{document}